\documentclass[letterpaper, 10 pt, conference]{ieeeconf}  

\IEEEoverridecommandlockouts                              

\usepackage{graphics} 
\usepackage{epsfig} 
\usepackage{times} 
\usepackage{amsmath, amsfonts} 
\usepackage{amssymb}  
\usepackage{cite}
\usepackage{subfig}
\usepackage{color, colortbl}

\usepackage{hyperref}  
\usepackage[font=small]{caption}

\title{SE(3) Linear Parameter Varying Dynamical Systems \\ for Globally Asymptotically Stable End-Effector Control}

\author{Sunan Sun$^*$ and Nadia Figueroa
\thanks{All authors are with the Department of Mechanical Engineering, University of Pennsylvania, Philadelphia PA 19104, USA}%
\thanks{$^*$Corresponding author. (e-mail: sunan@seas.upenn.edu)}
}

\begin{document}

\maketitle
\thispagestyle{empty}
\pagestyle{empty}

\begin{abstract}


Linear Parameter Varying Dynamical Systems (LPV-DS) encode trajectories into an autonomous first-order DS that enables reactive responses to perturbations, while ensuring globally asymptotic stability at the target. However, the current LPV-DS framework is established on Euclidean data only and has not been applicable to broader robotic applications requiring pose control. In this paper we present an extension to the current LPV-DS framework, named Quaternion-DS, which efficiently learns a DS-based motion policy for orientation. Leveraging techniques from differential geometry and Riemannian statistics, our approach properly handles the non-Euclidean orientation data in quaternion space, enabling the integration with positional control, namely SE(3) LPV-DS, so that the synergistic behaviour within the full SE(3) pose is preserved. Through simulation and real robot experiments, we validate our method, demonstrating its ability to efficiently and accurately reproduce the original SE(3) trajectory while exhibiting strong robustness to perturbations in task space.
\end{abstract}

\section{INTRODUCTION}\label{sec:intro}
 Adaptivity to new tasks and robustness to unforeseen perturbations and uncertainties are fundamental for safe integration of robots in human workspaces. On one hand, trajectory planning for position in $\mathbb{R}^d$ has significantly advanced and evolved into a multitude of approaches for different applications. These range from traditional path planning algorithms assuming a known environment and robot dynamics~\cite{UDE1993113, Mobilerobotsat, ALEOTTI2006409}, to more adaptive and reactive alternatives using Dynamical System (DS) or Dynamic Movement Primitives (DMP) to encode complex trajectories~\cite{TEXTBOOK, DMP1, DMP2}. On the other hand, trajectory planning for orientation in $\mathbb{S}^3$ requires more complex mathematical strategies due to the non-Euclidean nature of the orientation manifold and its multitude of representations such as Euler angles, rotation matrices and quaternions ~\cite{ORI_REVIEW}. 

For many tasks it might be sufficient to control solely the position of the end-effector of a robot, however, a truly dexterous manipulator requires full end-effector pose control. While orientation controllers do exist, they often require hard-wiring and re-programming of every new task, not allowing the adaptability, flexibility and robustness to perturbations achieved by reactive DS position control strategies. 
Such discrepancy in development has hindered the capability of a fully interactive robot. In practice, the decoupling becomes common where the position and orientation are computed and executed via two independent controllers. The cost, however, is the loss of the dependency between position and orientation inherent to any task, let alone the inflexibility that comes with a non-adaptive orientation controller.

\begin{figure}[!t]
\centering
\includegraphics[width=1\linewidth]{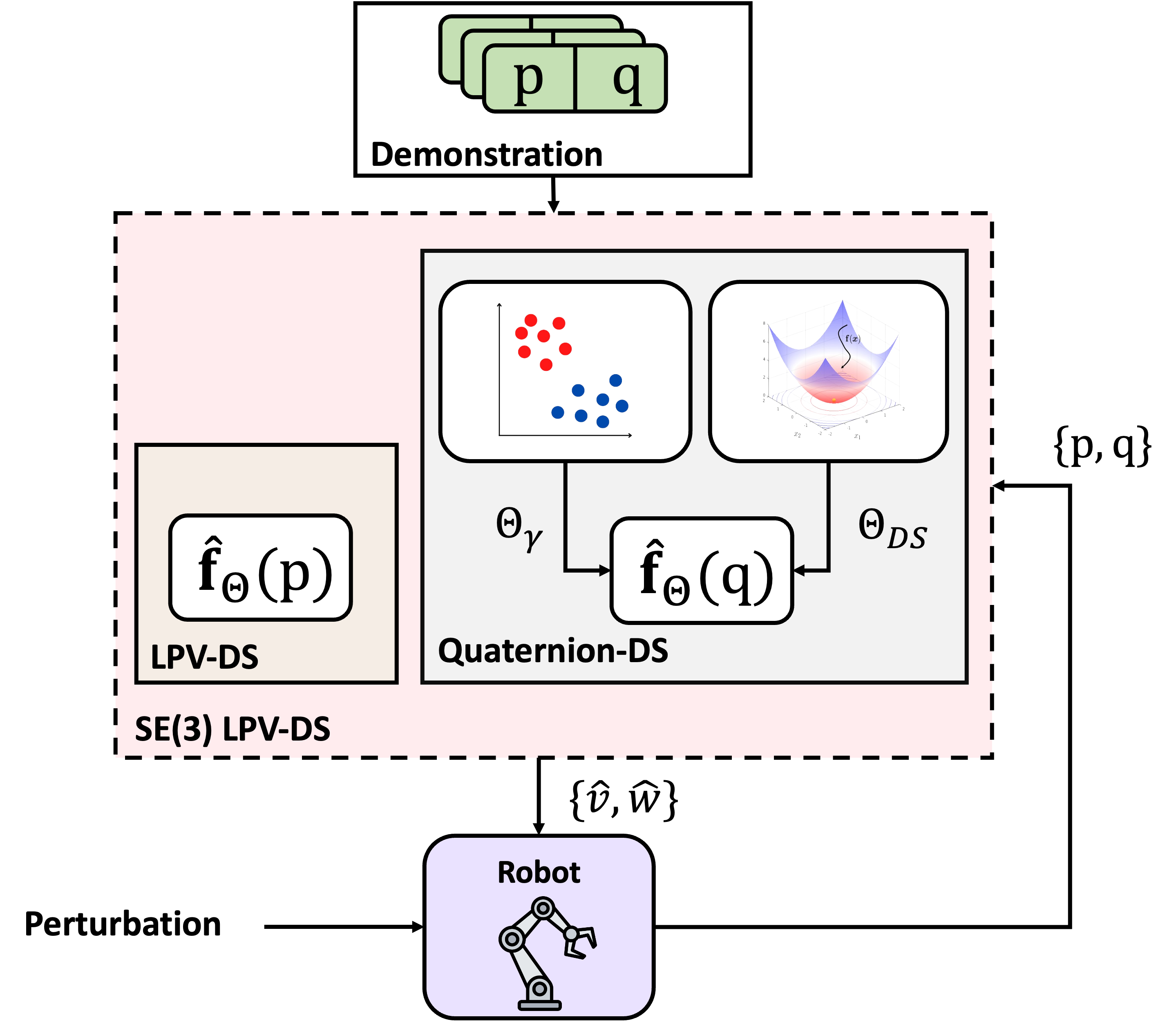}%
\hfill
\caption{The schematic of the \textbf{SE(3) LPV-DS} formulation which is composed of an ordinary LPV-DS \cite{PC-GMM} for position control and \textbf{Quaternion-DS} for orientation control; the architecture of Quaternion-DS consists of the clustering of the orientation trajectory and the optimization to minimize the prediction error; the resulting SE(3) LPV-DS takes the end-effector states: position $\bold{p}$ and orientation $\bold{q}$ as inputs, and generates the estimated desired linear velocity $\Hat{v}$ and angular velocity $\Hat{\omega}$, which are then passed down to command the robot via a low-level feedback controller.}
\label{fig:pipeline}
\vspace{-15pt}
\end{figure}


Encoding trajectories as Dynamical System (DS) has been increasingly popular for enabling adaptive behaviour in robotic control \cite{TEXTBOOK}. DS-based motion policies leverage redundancy of solutions in dynamic environments and embed an infinite set of feasible solutions in a single control law to overcome external uncertainties and perturbations~\cite{SEDS}.
Among existing learning frameworks, recent neural network (NN) based formulations for stable DS motion policies show promising results in encoding highly non-linear trajectories; as adopting normalizing flows~\cite{9341035}, euclideanizing flows~\cite{pmlr-v120-rana20a} or via contrastive learning~\cite{10214439}; however, most works are mainly focused on Euclidean data and designing a NN on orientation manifold is not easily feasible due to its black box nature~\cite{NN-ORI}. A recent work has proposed a diffeomorphism-based approach to learn stable motion policies on different Riemannian manifolds, including $\mathbb{R}^3 \times \mathbb{S}^3$ \cite{zhang2022learning}. Yet, diffeomorphic and NN based approaches, while expressive, are computational and sample inefficient, as well as incapable of adapting to new task parameters or environments.

The Linear Parameter Varying Dynamical System (LPV-DS) formulation, on the other hand, is the seminal framework in learning stable, time-independent DS-based motion policies from limited demonstrations~\cite{PC-GMM, TEXTBOOK}. As opposed to NN-based learning that requires many trajectories and substantial computation time to reach stable solutions, LPV-DS is effective in learning trajectory behaviour with minimal data and higher computational efficiency for potential real-time incremental learning, as showcased in a recent extension capable of learning LPV-DS motion policies in seconds \cite{sun2024directionality}.

In addition, LPV-DS is comprised of a statistical model --- a Gaussian Mixture Model (GMM) and a semi-definite optimization, making it an \textit{explainable} pipeline. This enables greater flexibility in engineering specific responses~\cite{LOCALLY_ACTIVE}, and allows for stability conditions constructed as constraints in the learning, offering a closed-form analytical solution to trajectory planning with theoretical guarantees such as stability and convergence. Furthermore, the LPV-DS framework grounded on trajectory data has shown to be able to generalize to new task instances in the Elastic-DS formulation \cite{li2023task}.

In this paper we extend the current LPV-DS framework to encode SE(3) trajectories, referred to as \textbf{SE(3) LPV-DS}. In the light of the LPV-DS framework, we first introduce the \textbf{Quaternion-DS} for adaptive orientation control. With the proper handling and techniques from Riemannian statistics and differential geometry, the Quaternion-DS framework can generate a stable rotational motion policy given the orientation trajectories. We then formulate a comprehensive SE(3) LPV-DS framework by integrating the Quaternion-DS with the established LPV-DS method, enabling full SE(3) pose control using the translation and quaternion representation instead of the homogeneous matrix representation. We evaluate our approach through extensive empirical validation on real robot experiments. Our results demonstrate that the SE(3) LPV-DS maintains the intrinsic relationship between position and orientation, while preserving all the perks of LPV-DS including global asymptotic stability, strong robustness to external perturbations, and high computational efficiency.

\textbf{Paper Organization:} Section~\ref{sec:prelim} introduces the mathematical preliminaries of the LPV-DS framework, Quaternion arithmetic and Riemannian statistics. The proposed methods are detailed in Section~\ref{sec:quat-ds} and evaluated in Section~\ref{sec:eval}. Source code for the work is available at \url{https://github.com/SunannnSun/quaternion_ds}.

\section{MATHEMATICAL PRELIMINARIES}
\label{sec:prelim}

\subsection{LPV-DS formulation}
\label{sec:lpvds}
We begin by introducing the existing LPV-DS framework for Euclidean data~\cite{PC-GMM}, which is typically used to encode position trajectories represented by the brown block in Figure~\ref{fig:pipeline}. This will serve as the foundation for the formulation of the Quaternion-DS.

Let $\xi, \dot{\xi} \in \mathbb{R}^d$ represent the kinematic robot state and velocity vectors. In the DS-based motion policy literature \cite{TEXTBOOK}, $\dot{\xi} = f(\xi)$ is a first-order DS that describes a motion policy in the robot's state space $\mathbb{R}^d$. The goal of DS-based learning from demonstration (LfD) is to infer $f(\xi): \mathbb{R}^d \rightarrow \mathbb{R}^d$ from data, such that any point $\xi$ in the state space leads to a stable attractor $\xi^*\in \mathbb{R}^d$, with $f(\xi)$ described by a set of parameters $\Theta$ and attractor $\xi^*\in\mathbb{R}^d$; mathematically $\dot{\xi} = f(\xi; \Theta, \xi^*) \Rightarrow \lim_{t \to \infty} \|\xi - \xi^*\| = 0$, i.e., the DS is globally asymptotically stable (GAS)~\cite{Khalil:1173048}.

Learning $\Dot{\xi}=f(\xi)$ can be framed as a regression problem, where the inputs are the state variables $\xi$ and the outputs are the first-order time derivative $\Dot{\xi}$. Such formulation gives rise to the utilization of statistical methods for estimating the parameters $\Theta$. However, standard regression techniques cannot ensure globally asymptotic stability. To alleviate this, the LPV-DS approach was first introduced in the seminal work of \cite{SEDS} as a constrained Gaussian Mixture Regression (GMR) and then formalized as the untied GMM-based LPV-DS approach in \cite{PC-GMM}, where a nonlinear DS is encoded as a mixture of continuous linear time-invariant (LTI) systems: 
\begin{equation} \label{eq:lpv_ds}
\begin{aligned}
    &\quad \quad \dot{\xi}=f(\xi; \Theta)=\sum_{k=1}^K \gamma_k(\xi)\left(\bold{A}_k \xi+b_k\right)\\
    \text{s.t.}&\,\,  \left\{\begin{array}{l}
    \left(\bold{A}_k\right)^T \bold{P}+\bold{P} \bold{A}_k=\bold{Q}_k, \bold{Q}_k=\left(\bold{Q}_k\right)^T \prec 0 \\
    b_k=-\bold{A}_k \xi^* \end{array}\right.
\end{aligned}
\end{equation}
where $\gamma_k(\xi)$ is the state-dependent mixing function that quantifies the weight of each LTI system $(\bold{A}_k \xi + b_k)$ and $\Theta = \{\theta_\gamma\}_{\gamma=1}^K = \{\gamma_k, \bold{A}_k, b_k\}_{k=1}^K$ is the set of parameters to learn. The constraints of the Eq. \ref{eq:lpv_ds} enforce GAS of the result DS derived from a parametrized Lyapunov function $V(\xi) = (\xi-\xi^*)^T\bold{P}(\xi-\xi^*)$ with $\bold{P}=\bold{P}^T\succ0$~\cite{PC-GMM, TEXTBOOK}.

To ensure GAS of Eq. \ref{eq:lpv_ds}, besides enforcing the Lyapunov stability constraints on the LTI parameters one must ensure that $0<\gamma_k(\xi)< 1$ and $\sum_{k=1}^K \gamma_k(\xi)= 1 ~\forall \xi \in \mathbb{R}^d$. As noted in \cite{PC-GMM}, this is achieved by formulating $ \gamma_k({\xi}) = \frac{\pi_k \mathcal{N}({\xi}| \theta_k)}{\sum_{j=1}\pi_j \mathcal{N}(\xi| \theta_j)}$ as the \textit{a posteriori probability} of the state $\xi$ from a GMM used to partition the nonlinear DS into linear components. Here, $K$ is the number of components corresponding to the number of LTIs, $\mathcal{N}({\xi}| \theta_k)$ is the probability of observing ${\xi}$ from the $k$-th Gaussian component parametrized by mean and covariance matrix $\theta_k =\{\mu_k,\bold{\Sigma}_k \}$, and $\pi_k$ is the prior probability of an observation from this particular component satisfying $\sum_{k=1}^K \pi_k = 1$.

In \cite{PC-GMM} a two-step estimation framework was proposed to estimate the GMM parameters $\Theta_{\gamma}=\{\pi_k, \mu_k, \bold{\Sigma}_k\}_{k=1}^K$ and the DS parameters $\Theta_{DS} = \{\bold{A}_k, b_k\}_{k=1}^K$ forming $\Theta=\{\Theta_{\gamma}, \Theta_{DS}\}$. First, given the set of reference trajectories $\mathcal{D}:=\{\xi^{\mathrm{ref}}_i\ \dot{\xi}^{\mathrm{ref}}_i\}_{i = 1}^N$, where $i$ is the sequence order of the sampled states, a GMM is fit to the position variables of the reference trajectory, $\{\xi^{ref}_i\ \}_{i = 1}^N$,  to obtain $\Theta_{\gamma}$. The optimal number of Gaussians $K$ and their placement can be estimated by model selection via Expectation-Maximization or via Bayesian non-parametric estimation. Then, $\Theta_{DS}$ are learned through a semi-definite program minimizing reproduction accuracy subject to stability constraints \cite{PC-GMM,TEXTBOOK}.


\subsection{Quaternion Arithmetic}
The unit quaternion is a compact representation of orientation defined by $\bold{q} = w + x\bold{i} + y \bold{j} + z \bold{k} \in \mathbb{H} \subset \mathbb{R}^4$ with unit norm $\| \bold{q} \| = 1$. The inverse orientation is represented by its conjugate $\Bar{\bold{q}} = w - x\bold{i} - y \bold{j} - z \bold{k}$. Both quaternions $\bold{q}$ and $-\bold{q} = -w - x\bold{i} - y \bold{j} - z \bold{k}$ represent the same orientation.

Given two quaternions $\bold{q}_1$ and $\bold{q}_2$, we can compute the displacement as $\Delta \bold{q} = \Bar{\bold{q}}_1 \circ \bold{q}_2$ which is the quaternion product of $\Bar{\bold{q}}_1$ and $\bold{q}_2$. The quaternion multiplication is not commutative; i.e. the order of multiplication matters and changing order will result in a different outcome. For example, the $\Delta \bold{q}$ computed above represents the amount of rotation required to rotate $\bold{q}_1$ onto $\bold{q}_2$ with respect to the body frame. If the time difference $dt$ is provided, we can therefore compute the angular velocity by first converting the displacement in axis-angle representation as follows: 
\begin{equation}    
\omega = \frac{2}{dt}\arccos{(w)} \frac{(x, y, z)}{\| (x, y, z)\|}
\end{equation}
Note again the computed $\omega$ is the angular velocity expressed in the body frame not the world frame due to the specific order of the quaternion multiplication.


\subsection{Riemannian Manifold} \label{sec:riem}
Unit quaternions can also be conceptualized as residing on a three-dimensional hypersphere with a radius of 1, known as the 3-sphere or $\mathbb{S}^3$, which is embedded in four-dimensional Euclidean space $\mathbb{R}^4$. Recognizing the 3-sphere is a Riemannian manifold allows us to employ techniques from differential geometry and Riemannian statistics. First of all, a Riemannian manifold is a smooth manifold equipped with positive definite inner product defined in the tangent space at each point. This metric allows for the measurement of distances, angles, and other geometric properties on the manifold. Hereafter, we only consider the unit sphere as pertaining to the unit quaternions, and the provided formulas are mostly specific to the unit sphere. For clarity, we denote elements of the manifold in bold and elements in tangent space in fraktur typeface; i.e. $\bold{p} \in \mathcal{M}$ and $\frak{q} \in T_\bold{p}\mathcal{M}$. 

The notion of distance on the Riemannian manifold is a generalization of straight lines in Euclidean spaces. The minimum distance paths that lie on the curve, also called geodesics, are defined as $d(\bold{p}, \bold{q}) = \arccos(\bold{p}^T\bold{q})$ between two points on unit sphere, or $\bold{p}, \bold{q}\in \mathbb{S}^d$~\cite{RIEM-GEO, RIEM-GEO2}. We can also compute the Riemannian equivalent of mean and covariance as follows,
\begin{equation} \label{eq:riem_mu_sigma}
\begin{aligned}
            &\quad \Tilde{\mu} = \operatornamewithlimits{argmin}_{\bold{p} \in \mathbb{S}^{d}} \sum_{i=1} ^N  d(\bold{q}_i,\ \bold{p})^2\\
            \Tilde{\bold{\Sigma}}&= \frac{1}{(N-1)} \sum_{i=1}^N \log_{\Tilde{\mu}}(\bold{p}_i)\log_{\Tilde{\mu}}(\bold{p}_i)^T.
\end{aligned}
\end{equation}
The average $\Tilde{\mu}$, defined as the center of mass on unit sphere, employs the notion of the Fréchet mean~\cite{RIEM-MEAN}, which extends the sample mean from $\mathbb{R}^d$ to Riemannian manifolds $\mathcal{M}$. In practice, $\Tilde{\mu}$ can be efficiently computed in an iterative approach~\cite{RIEM-COV}. The empirical covariance $\Tilde{\bold{\Sigma}}$ captures the dispersion of data in tangent space $T_\bold{p}\mathcal{M}$, where the logarithmic map $\log_\bold{p}:\mathcal{M} \rightarrow  T_\bold{p}\mathcal{M}$ maps a point on the Riemannian manifold to the tangent space defined by the point of tangency $\bold{p}$~\cite{RIEM-COV, RIEM-STAT, RIEM-STAT-2}:
\begin{equation} \label{eq:riem_log}
    \frak{q} = \log_\bold{p}(\bold{q}) = d(\bold{p}, \bold{q}) \frac{\bold{q} - \bold{p}^T\bold{q}\bold{p}}{\|\bold{q} - \bold{p}^T\bold{q}\bold{p}\|}.
\end{equation}
The inverse map is the exponential map $\exp_\bold{p}: T_\bold{p}\mathcal{M} \rightarrow \mathcal{M}$ which maps a point in tangent space of $\bold{p}$ to the manifold so that the mapped point lies in the direction of the geodesic starting at $\bold{p}$~\cite{RIEM-COV, RIEM-STAT, RIEM-STAT-2}:
\begin{equation} \label{eq:riem_exp}
    \bold{q} = \exp_\bold{p}(\frak{p}) = \bold{p}\cos(\| \frak{p} \|) + \frac{\frak{p}}{\| \frak{p} \|}\sin(\| \frak{p} \|).
\end{equation}

\begin{figure}[!t]
\centering
\includegraphics[width=1\linewidth]{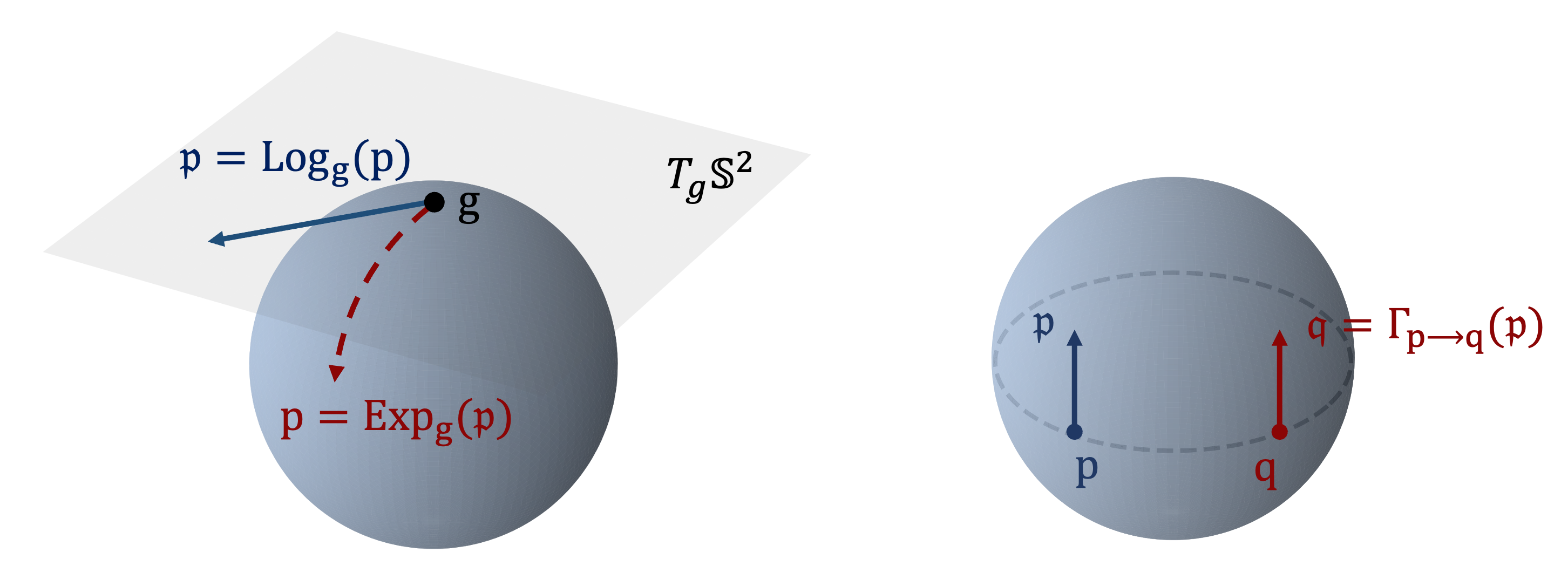}%
\hfill
\caption{Illustrative examples of the operations on Riemannian geometry: exponential/logarithmic mapping (left) and parallel transport (right), which are depicted on a $\mathbb{S}^2$ manifold embedded in $\mathbb{R}^3$ }
\label{fig:riem}
\vspace{-10pt}
\end{figure}

The parallel transport $\Gamma_{\bold{p} \rightarrow \bold{q}}(\frak{p}): T_\bold{p}\mathcal{M} \rightarrow T_\bold{q}\mathcal{M}$ transports a vector $\frak{p}$ in the tangent space $T_\bold{p}\mathcal{M}$ at a point $\bold{p}$ on a manifold $\mathcal{M}$ to a vector in the tangent space $T_\bold{q}\mathcal{M}$ at point $\bold{q}$, along geodesic:~\cite{do1992riemannian}
\begin{equation} \label{eq:riem_pt}
    \Gamma_{\bold{p} \rightarrow \bold{q}}(\frak{p}) = \frak{p} - \frac{\log_\bold{p}(\bold{q})^T \frak{p}}{d(\bold{p}, \bold{q})^2} (\log_\bold{p}(\bold{q}) + \log_\bold{q}(\bold{p})).
\end{equation}

The L2 norm $\| \log_{\bold{p}} (\bold{q}) \|_2$ in the tangent space $T\mathbb{S}^d$ is equal to the geodesic distance between points $\bold{p}$ and $\bold{q}$ on the manifold: $d(\bold{p}, \bold{q})$. This provides an effective measure of deviations from a single point, allowing us to construct Gaussians on the manifold, as will be shown in Section~\ref{sec:gmm}. However, this is true only when $\bold{p}$ is the point of tangency. In general, the inner product between two vectors in $T\mathbb{S}^d$ is not equal to the geodesic distance between their corresponding points in $\mathbb{S}^d$.~\cite{do1992riemannian}

\section{QUATERNION DYNAMICAL SYSTEM}
\label{sec:quat-ds}

Section~\ref{sec:lpvds} introduces the general framework of LPV-DS that is only suitable to the Euclidean data such as position trajectories. Note that a learned DS via LPV-DS takes the form of $\dot{\xi} = f(\xi)$, and when the system is GAS, $\xi$ will reach the target and $\dot{\xi}$ will diminish to zero over time. When it comes to orientation, we expect that the learned DS might take different forms with different inputs and outputs, but the objective should remain the same: i) the system can reach the target and ii) the system is stable at the target.

Among the various representations of orientations, quaternion stands out as compact and singularity-free. In accordance with the Euclidean LPV-DS, it is most intuitive to construct its orientation counterpart by taking quaternion as input and generating either angular velocity or the derivative of quaternion as output. However, neither angular velocity nor derivative of quaternion lives in quaternion space, meaning no direct affine mapping from quaternions to any of them.

Say we have a trajectory of unit quaternion, $\mathcal{Q} := \{ \bold{q}^i\}_{i=1}^N$ and $\bold{q}_{att}$ is denoted as the attractor or target of this quaternion trajectory, we first project all the elements in the quaternion trajectory to the tangent space defined by $\bold{q}_{att}$ via the logarithmic map in Eq.~\ref{eq:riem_log}, 
\begin{align} \label{eq:q_att}
     \mathfrak{q}_{att}^i = \log_{\bold{q}_{att}}{(\bold{q}^i}) \quad \forall\ i=1,\ \dots, \ N
\end{align}
where all the unit quaternions $\bold{q}^i \in \mathbb{S}^3$ have now been projected onto the tangent space defined by the attractor, $\mathfrak{q}_{att}^i \in T_{\bold{q}_{att}}\mathbb{S}^{3}$. Each projected vector represents the direction and distance from $\bold{q}_{att}$ to $\bold{q}^i$ along a geodesic path on the manifold while preserving the Riemannian metric of two vectors on the Riemannian manifold. We note the analogy of Eq.~\ref{eq:q_att} to its Euclidean equivalent $(\xi - \xi^*)$ in Section~\ref{sec:lpvds}.

As we argue that neither angular velocity $\omega$ nor the derivative of quaternion $\Dot{q}$ lives in quaternion space, we resolve to a discrete system that outputs the next desired orientation. Provided the angular velocity and time difference, we can integrate forward and compute:
\begin{equation}
    (\bold{q}^i)^{des} = \bold{q}^i \circ (\omega^i \times dt),
\end{equation}
where $(\omega^i \times dt)$ should be converted to the quaternion representation before being composed with the current quaternion. We then perform logarithmic map in Eq.~\ref{eq:riem_log}, 
but this time w.r.t. the corresponding current state:
\begin{equation}\label{eq:q_body}
     (\mathfrak{q}_{body}^i)^{des} = \log_{\bold{q}^{i}}{(\bold{q}^i)^{des}} \quad \forall\ i=1,\ \dots, \ N.
\end{equation}
Via Eq.~\ref{eq:q_body} obtain a new set of vectors representing the distance or displacement from every orientation to their desired state, and each vector resides in a unique tangent space defined by the current state $\bold{q}^i$. One can also draw the analogy between Eq.~\ref{eq:q_body} and the linear velocity $\Dot{\xi}$ in Section~\ref{sec:lpvds}. When the system is approaching the target, $\Dot{\xi}$ should point towards the target and reach zero asymptotically. The discrete counterpart of velocity in quaternion space, or $\mathfrak{q}_{body}^i$ embodies the same idea as the displacement should lead to the target and gradually reduce to zero.

For mathematical completeness, we perform an additional step by parallel transporting each individual vector $\mathfrak{q}_{body}^i$ from their current state to the attractor as in Eq.~\ref{eq:riem_pt}:
\begin{equation}
     (\mathfrak{q}_{att}^i)^{des}= \Gamma_{\bold{q}^i \rightarrow \bold{q}_{att}}(\mathfrak{q}_{body}^i)^{des}
\end{equation}
so that both $\mathfrak{q}_{att}^i$, the vector from the attractor $\rightarrow$ current state, and $(\mathfrak{q}_{body}^i)^{des}$, the vector from the current state $\rightarrow$ desired state, are expressed in the same tangent space defined by the quaternion attractor $\bold{q}_{att}$.

Putting things together the quaternion-DS should bear the following form according to the LPV-DS framework in Eq.~\ref{eq:lpv_ds},
\begin{equation} \label{eq:quat_ds}
\begin{aligned}
    &(\Hat{\mathfrak{q}}_{att}^i)^{des} = \sum_{k=1}^K \gamma_k(\bold{q}^i) \bold{A}_k\log_{\bold{q}_{att}} \bold{q}^i,\\
    \text{s.t.}\,\,  &\left\{\begin{array}{l}
    \sum_{j=1}^K \sum_{k=1}^K \gamma_j(\bold{q}^i) \gamma_k(\bold{q}^i) \bold{A}_j^T \bold{P } \bold{A}_k -  \bold{P}  \prec 0 
    \end{array}\right.
\end{aligned}
\end{equation}
where $(\Hat{\mathfrak{q}}_{att}^i)^{des}$ is the estimated desired orientation output by the DS, $\gamma_{*}(\bold{q}^i)$ is the state-dependent mixing function that quantifies the weight of each LTI system, $K$ is the number of LTI systems partitioned by the statistical model, and $\bold{A} \in \mathbb{R}^{4 \times 4}$ is the linear parameters of each LTI system. The constraints enforce the globally asymptotic stability (GAS) at the target, which is derived from the Lyapunov stability conditions for a discrete system.


In the following sections, we will detail a) the stability analysis of the quaternion-DS, and how to b) convert the output of quaternion-DS to angular velocity for robot control, c) obtain the mixing functions $\gamma_{*}(\cdot)$ via statistical model, and d) learn the linear parameters $\bold{A}$ via optimization.
\subsection{Stability analysis} \label{sec:stability}
To derive the Lyapunov constraints that enforce the GAS of our quaternion-DS as in Eq.~\ref{eq:quat_ds}, we choose the Lyapunov function to be quadratic in the tangent space defined by the attractor with $\bold{P}  = \bold{P}^T \succ 0$ as, 
\begin{align} \label{eq:lyap_func}
    &V(\bold{q}^i, \bold{q}_{att}) = V(\mathfrak{q}^i_{q_{att}}) =\mathfrak{q}_{q^i_{att}}^T \bold{P} \mathfrak{q}_{q^i_{att}} > 0, 
\end{align} 
Observe that for all $\bold{q}^i$ that is not $\bold{q}_{att}$, the Lyapunov function in Eq.~\ref{eq:lyap_func} is always positive. According to the Lyapunov stability theory for discrete systems~\cite{Lyapunov_Discrete}, the Lyapunov function must satisfy that its difference is always negative and only equal to zero at the target equilibrium:
\begin{equation} \label{eq:lyap_deriv}
\begin{aligned}   
    \Delta V(\bold{q}^i) & = V(\bold{q}^{i+1}) - V(\bold{q}^i)\\
    & = \mathfrak{q}_{q^{i+1}_{att}}^T \bold{P} \mathfrak{q}_{q^{i+1}_{att}} - \mathfrak{q}_{q^i_{att}}^T \bold{P} \mathfrak{q}_{q^i_{att}} \\
    & = \mathfrak{q}_{q^i_{att}}^T \left(\sum_{j=1}^K \sum_{k=1}^K \underbrace{\gamma_j(\bold{q}^i) \gamma_k(\bold{q}^i)}_{\succ 0} \bold{A}_j^T \bold{P } \bold{A}_k -  \bold{P} \right)\mathfrak{q}_{q^i_{att}}\\
\end{aligned}
\end{equation}
Given that the mixing function $\gamma_{*}(\cdot)$ being positive and the matrix $\bold{P}$ being positive definite, we have to ensure that the summation of every terms within the parenthesis in Eq.~\ref{eq:lyap_deriv} is negative definite. We note that when enforcing the linear parameter $\bold{A}_k$ of each LTI system to be negative definite: $\bold{A}_k \prec 0 \quad \forall k = 1, \ \dots \ , K $, we are ensuring that
\begin{equation}
\begin{aligned}
        \bold{A}_k^T \bold{P} \bold{A}_k &\prec 0 \quad \forall k = j \\
        \bold{A}_j^T \bold{P} \bold{A}_k &\preceq 0 \quad \forall k \neq j,
\end{aligned}
\end{equation}
and hence the difference in Eq.~\ref{eq:lyap_deriv} is always negative except at the equilibrium.


When $\bold{q}^i$ reaches the target equilibrium $\bold{q}_{att}$, we have the Lyapunov function and its difference equal to zero,
\begin{equation}
\begin{aligned}
         &V(\bold{q}_{att}, \bold{q}_{att}) = \bold{0}^T \bold{P} \bold{0} = 0\\ 
         \Delta V(\bold{q}_{att}) 
         &= \bold{0}^T \left(\sum_{j=1}^K \sum_{k=1}^K \gamma_j(\bold{q}^i) \gamma_k(\bold{q}^i) \bold{A}_j^T \bold{P } \bold{A}_k -  \bold{P} \right)\bold{0} \\
         & = 0. 
\end{aligned}
\end{equation}
Hence, the Lyapunov function and its time difference satisfy all the necessary conditions to ensure the GAS of the quaternion-DS in the Lyapunov sense. 

\subsection{Conversion to Angular Velocity for Control}
In a typical robotic application, the estimated desired state from the quaternion-DS formulation cannot be directly used by the low-level controller, and we need to recover the time derivative of the state space, that is the angular velocity in this case. Provided that a time difference $dt$ is known, we first parallel transport the estimated $(\Hat{\mathfrak{q}}_{att}^i)^{des}$ from the attractor back to the current state $\bold{q}^i$, 
\begin{equation}
        (\Hat{\mathfrak{q}}_{body}^i)^{des}  = \Gamma_{\bold{q}_{att} \rightarrow \bold{q}^i}(\Hat{\mathfrak{q}}_{att}^i)^{des},
\end{equation}
where the new vector is the estimated desired displacement expressed in the body frame. We then map this vector from the tangent space back to the quaternion space, 
\begin{equation}
        (\Hat{\bold{q}}^i)^{des} = \exp_{\bold{q}^i}{(\Hat{\mathfrak{q}}_{body}^i)^{des}},
\end{equation}
where the vector is the estimated desired orientation in quaternion space; and lastly compute the angular velocity given the current state and time difference,
\begin{equation}
        \Hat{\omega}^i = (\Bar{\bold{q}}^i \circ (\Hat{\bold{q}}^i)^{des}) / dt.
\end{equation}
The result of quaternion product needs to be converted into axis-angle representation before dividing it by $dt$ to generate the proper angular velocity. Due to the specific order of the quaternion multiplication above, the computed $\omega$ is the estimated angular velocity w.r.t. the body frame, not the fixed frame. This $\omega$ can then be passed down to a low level controller such as a twist controller for a robot as in Fig.~\ref{fig:pipeline}.

\subsection{Quaternion Mixture Model} \label{sec:gmm}

One advantage of LPV-DS is the use of statistical model to capture the intrinsic structure of trajectory data while accounting for the uncertainty within trajectory. This is manifested as the mixing function $\gamma(\cdot)$ in Eq.~\ref{eq:quat_ds}, which quantifies the weight of each LTI system and governs the transition between each LTI system at a given state.   

Using GMM to cluster Euclidean data is straightforward as introduced in Section~\ref{sec:lpvds} where one can choose variants of GMM and inference methods to construct the mixing function. However, clustering quaternions using GMM is challenging because GMM assumes Euclidean metrics while quaternions are non-Euclidean. However, by recognizing that the unit quaternions reside on the Riemannian manifold or the 3-sphere, we can employ the techniques from differential geometry and apply the same treatment as in the previous Section~\ref{sec:stability} --- projecting quaternion data onto the tangent space, locally approximating them as Euclidean vectors and fitting a GMM to the tangent vector. But caution is needed and we argue that the following quaternion mixture model is only useful and specific to our application and may not serve as a general framework for clustering sparse quaternions.

The orientation trajectory has a special trait that is missing from sparse quaternion data, that is the trajectory has an end point. Projecting unit quaternion data w.r.t. this end point bears physical meanings as we discussed in Section~\ref{sec:quat-ds}; those tangent vectors represent the displacement from the target to the current state in terms of orientation. By clustering these tangent vectors in the tangent space, we are also implicitly segmenting the unit quaternion data in quaternion space. We argue that the tangent vectors that share similar displacement from the attractor should also be close to each other in the quaternion space, and this is only true when the tangent vector preserves the intrinsic geometry of the trajectory and Euclidean metrics is a fair approximation. 

\begin{figure}[!t]
\centering
\includegraphics[width=1\linewidth]{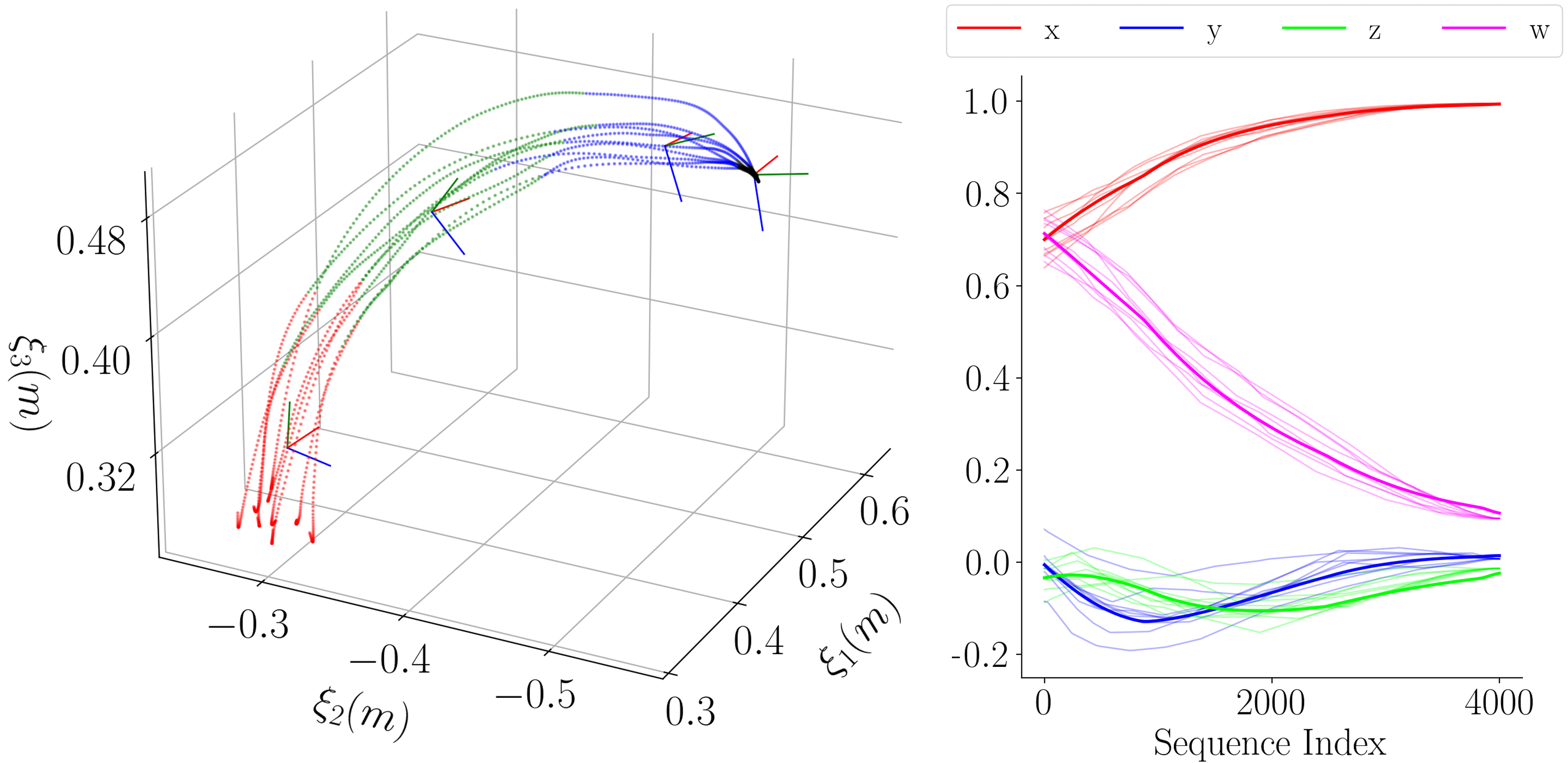}%
\hfill
\vspace{-5pt}
\caption{The quaternion mixture model ($K=4$) with the mean orientation of each Gaussian (left); note the color is only indicative of the clustering results on the quaternion trajectory, not position. The simulation result of quaternion-DS (right) shows the evolution of the quaternions in its 4 coordinates, where the thin lines are demonstration trajectory, and the thick lines are reproduction.}
\label{fig:quat_ds}
\vspace{-15pt}
\end{figure}

Another assumption underlying the LPV-DS framework is that trajectory should not \textit{self-intersect}, given that an autonomous first-order DS system cannot handle higher-order derivatives nor sequential information. The constraint holds across various benchmark dataset for LPV-DS framework~\cite{SEDS, PC-GMM}, and is easy to enforce when data is collected in a continuous manner, as opposed to sparse points. Hence, this is a fair assumption for orientation data as well; i.e. no orientation trajectory should reach the same orientation more than once before stopping at the target. Inadvertently, this constraint also prevents the scenarios where a trajectory contains both $\bold{q}$ and $-\bold{q}$ which represent same orientation in task space while are drastically distinct vectors in quaternion space and its tangent space.

Say we have a trajectory of unit quaternion, $\mathcal{Q} = \{ \bold{q}^i\}_{i=1}^N$, we project all its elements to the tangent space defined by $\bold{q}_{att}$ via logarithmic map in Eq.~\ref{eq:riem_log}, and obtain the set of tangent vectors $\mathfrak{q}_{att}^i $ as in Eq.~\ref{eq:q_att}. When a GMM is fit to the tangent vectors, we can retrieve a label for each orientation data. Grouping the data points with the same assignment, we can then compute the mean $\Tilde{\mu}$ and empirical covariance $\bold{\Tilde{\Sigma}}$ as in Eq.~\ref{eq:riem_mu_sigma}. When computing the probability of observing a particular unit quaternion on a Gaussian: 
\begin{equation}\label{eq:quat_gauss}
    \mathcal{N}(\log_{\Tilde{\mu}}(\bold{q}^i) \vert \mu = \bold{0} ,\ \Sigma = \bold{\Tilde{\Sigma}}),
\end{equation}
note that the unit quaternion $\bold{q}$ is projected via logarithmic map w.r.t. the mean $\Tilde{\mu}$, and the actual mean $\mu$ of the Gaussian is a zero vector as the logarithmic map of the mean w.r.t. itself is zero or the origin in tangent space.

We can then formulate the mixing function in Eq.~\ref{eq:quat_ds} as the \textit{a posteriori probability} of the quaternion $\bold{q}$ on GMM,
\begin{equation}\label{eq:quat_gamma}
     \gamma_k({\bold{q}}) = \frac{\pi_k \mathcal{N}({\frak{q}_{\Tilde{\mu}_k}}| \theta_k)}{\sum_{j=1}\pi_j \mathcal{N}( \frak{q}_{\Tilde{\mu}_j}\vert \theta_j)},
\end{equation}
where the quaternion is mapped w.r.t. the mean of each Gaussian component respectively, and the parameters of each Gaussian must contain a zero mean. Such formulation enforces that $0<\gamma_k(\bold{q})< 1$ and $\sum_{k=1}^K \gamma_k(\bold{q})= 1 ~\forall \bold{q} \in \mathbb{H}$, ensuring the GAS of Eq. \ref{eq:quat_ds} as discussed in Section~\ref{sec:stability}.


Note that the mixture model only partitions the original trajectory $\mathcal{Q} = \{ \bold{q}^i\}_{i=1}^N$ , leaving the negative trajectory $\{-\bold{q}^i\}_{i=1}^N$ in the other half of the quaternion space unmodeled. We can mirror each Gaussian components by inverting the means $\Tilde{\mu}_k$, hence doubling the Gaussian components and covering the entire quaternion space.

\subsection{Optimization}
For mathematical completeness, we present the semi-definite optimization formulation that learns the linear parameters $\bold{A}_k$ of each LTI system. Following the same principle in the previous sections, the optimization is performed in the tangent space by minimizing the L2 norm of the error between the reference desired orientation and the prediction,
\begin{equation}\label{eq:opt}
\begin{aligned}
& \underset{\Theta}{\text{minimize}}
& & \sum_{i=1}^N \left\| (\Hat{\mathfrak{q}}_{att}^i)^{des}  -   (\mathfrak{q}_{att}^i)^{des} \right\|^2 \\
& \text{subject to}
& & \bold{A}_k \prec 0, \quad \forall k = 1, \dots, K, \\
\end{aligned}
\end{equation}
where the reference desired orientation is computed by Eq.~\ref{eq:q_att}, the prediction is computed by Eq.~\ref{eq:quat_ds} and Eq.~\ref{eq:quat_gamma}, and the constraint comes from the stability analysis in Section~\ref{sec:stability}. In Fig.~\ref{fig:quat_ds}, we illustrate a learned quaternion-DS. 

\section{SE(3) LPV-DS for Pose Control} \label{sec:se3}
While Quaternion-DS offers a comprehensive framework for encoding orientation trajectory and generating rotational motion policy in quaternion space, running Quaternion-DS with any Euclidean LPV-DS concurrently does not solve the missing synergy within the pose as discussed in Section~\ref{sec:intro}. 

\begin{figure}[!t]
\centering
\includegraphics[width=0.95\linewidth]{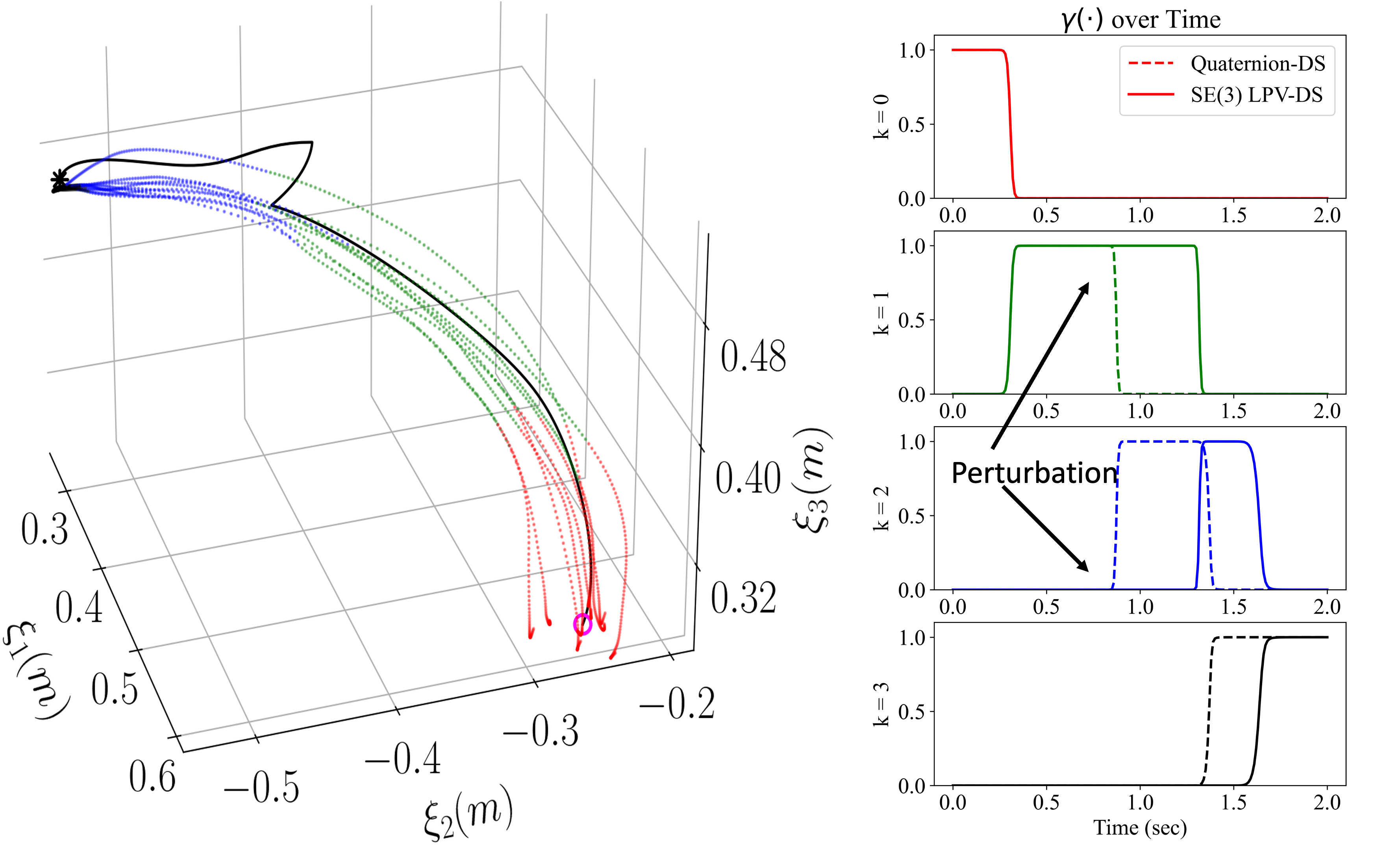}%
\hfill
\caption{The simulation results of SE(3) LPV-DS in the event of perturbation (left), where the demonstration data and clustering results are scatter points in color and the reproduction is the black curve; the value of mixing function $\gamma(\cdot)$ corresponding to each Gaussian ($K=4$) during the simulation (right), where the solid lines are computed in SE(3) LPV-DS by Eq.~\ref{eq:se3_gamma}, and the dashed lines are computed in Quaternion-DS by Eq.~\ref{eq:quat_gamma}.
}
\label{fig:perturb_ver}
\vspace{-10pt}
\end{figure}

SE(3) LPV-DS is an extension of the Quaternion-DS by introducing an implicit dependency between position and orientation. Previously, either in Euclidean LPV-DS or Quaternion-DS, the GMMs partition the trajectory in either $\mathbb{R}^3$ or $\mathbb{S}^3$, and the formulated mixing function $\gamma(\cdot)$ only accounts for the LTI systems in their respective vector space as discussed in Section~\ref{sec:lpvds} and Section~\ref{sec:gmm}. On the other hand, SE(3) LPV-DS concatenates the position and the projected orientation as an augmented input, $\left[ \bold{p},\  \frak{q}_{att}\right]^T \in \mathbb{R}^3 \times T\mathbb{S}^3$, and fits a GMM to the augmented input in the combined vector space. This is possible due to the construction of Quaternion-DS where the tangent vectors provide an effective measure in deviation w.r.t. the attractor by preserving the geodesic between points on the manifold when the point of tangency is the attractor. Similar to the Section~\ref{sec:gmm}, we can recover the label of each input data from the clustering results and construct the corresponding Gaussians with the probability as follows,

\begin{equation} \label{eq:se3_prob}
\begin{aligned}
        \mathcal{N}\bigg(\begin{bmatrix}
            \bold{p}\\
            \log_{\Tilde{\mu}}(\bold{q})
        \end{bmatrix} \bigg| \ 
        &\mu = \begin{bmatrix}
            \mu_{\bold{p}}\\
            \bold{0}
        \end{bmatrix}, \\
        &\Sigma = \frac{1}{N-1} \sum_{i=1}^N
        \begin{bmatrix}
        \bold{p} - \mu_{\bold{p}}\\
         \log_{\Tilde{\mu}}(\bold{q})
        \end{bmatrix}
        \begin{bmatrix}
        \bold{p} - \mu_{\bold{p}}\\
         \log_{\Tilde{\mu}}(\bold{q})
        \end{bmatrix}^T
        \bigg),    
\end{aligned}
\end{equation}
where the mean of the Gaussian is the positional mean concatenated with a zero vector, $\bold{0} \in \mathbb{R}^4$, and the deviation from mean is formulated by combining the  distance in Euclidean space $\bold{p} - \mu_{\bold{p}}$ and the geodesic on the manifold $\log_{\Tilde{\mu}}(\bold{q}_i)$. Note when computing the probability of a given pose, the quaternion component of the input needs to be transformed via logarithmic map w.r.t. the mean $\Tilde{\mu}$ while the position part of the state remains unchanged. This gives rise to the mixing function $\gamma(\cdot)$ with the following form,
\begin{equation}\label{eq:se3_gamma}
         \gamma_k\left(\begin{bmatrix}
            \bold{p}\\
            \bold{q}
        \end{bmatrix}\right) = \frac{\pi_k \mathcal{N}\bigg(\begin{bmatrix}
            \bold{p}\\
            \log_{\Tilde{\mu}_k}(\bold{q})
        \end{bmatrix}\bigg|\ \theta_k\bigg)}{\sum_{j=1}\pi_j \mathcal{N}\bigg( \begin{bmatrix}
            \bold{p}\\
            \log_{\Tilde{\mu}_j}(\bold{q})
        \end{bmatrix}\bigg|\ \theta_j\bigg)},
\end{equation}
where the parameters of each Gaussian $\theta$ should take the form of Eq.~\ref{eq:se3_prob}. This formulation of $\gamma(\cdot)$ allows the SE(3) LPV-DS to account for both position and orientation while partitioning the trajectory into Gaussian components and learning the linear parameters $\bold{A}$ during optimization. When reproducing the trajectory at any give state, the SE(3) LPV-DS assigns value to each $\gamma(\cdot)$ according to the pose rather than position or orientation only.

Fig.~\ref{fig:perturb_ver} presents a comparison result between the SE(3) LPV-DS and Quaternion-DS when responding to perturbation. The perturbation only occurs in position, and is manually added when a robotic system (e.g. end effector of a robot arm) is transitioning from the second LTI system (green) to the third LTI system (blue). We illustrate the response by plotting the value of mixing function $\gamma(\cdot)$ over time, which represents the weight each LTI system receives at a given state; e.g. a value of $\gamma_{k} = 1$ means that the robotic system is fully governed by the $k$-th LTI system. We observe that when the perturbation occurs at $t=0.8s$, the Quaternion-DS alone is completely unaware and proceeds to transition to the next LTI (blue). On the other hand, the SE(3) LPV-DS delays the transition and remains in its current LTI (green) until the position is rectified.

By fitting the mixture model to the pose trajectory and learning a single DS in the combined vector space, our approach coordinates position and orientation together in a coupled manner. By contrast, previous works, e.g. \textit{Neural Ordinary Differential Equation solvers} (NODEs)~\cite{sNODES} and \textit{contracting dynamical system primitives} (CDSP)~\cite{contraction}, learn separate autonomous nonlinear DS for position and orientation. To the best of our knowledge, SE(3) LPV-DS is the first DS-based motion policy that preserves the synergy inherent to any task while guaranteeing robustness to perturbations.


\section{EXPERIMENT}
\label{sec:eval}

\subsection{Benchmark Comparison}
\textbf{Baseline:} We compared our approach against the baseline method --- \textit{Neural Ordinary Differential Equation solvers (NODEs) which has been recently exploited to learn vector fields in trajectory-based learning~\cite{sNODES}}. With an equivalent parameter size, NODEs converges significantly faster than its neural-based predecessors, e.g. the \textit{Imitation Flow} (iFlow)~\cite{iFlow}, while still being able to accurately encode trajectory including both position and orientation.

\textbf{Dataset: }We evaluated our approach on the \textit{RoboTasks9} dataset from~\cite{sNODES}. The dataset of 9 real-world tasks includes box opening, plate stacking and etc., in which both the position and orientation of the robot's end-effector vary over time. Each task, provided kinsethetically by humans, contains 9 trajectories with a size of 1000 observations each including both position $\bold{p} \in \mathbb{R}^3$ and orientation $\bold{q} \in \mathbb{H}$.

\textbf{Metrics: }To assess the learning performance, we report the widely used metrics: \textit{Dynamic Time Warping error} (DTW) for end-effector position and \textit{Quaternion error}~\cite{QUAT_ERROR} for orientation. In addition, we measure the model complexity and computation time; both are crucial for continual and incremental learning in real-time scenario.

\begin{figure}[!t]
\centering
\includegraphics[width=0.9\linewidth]{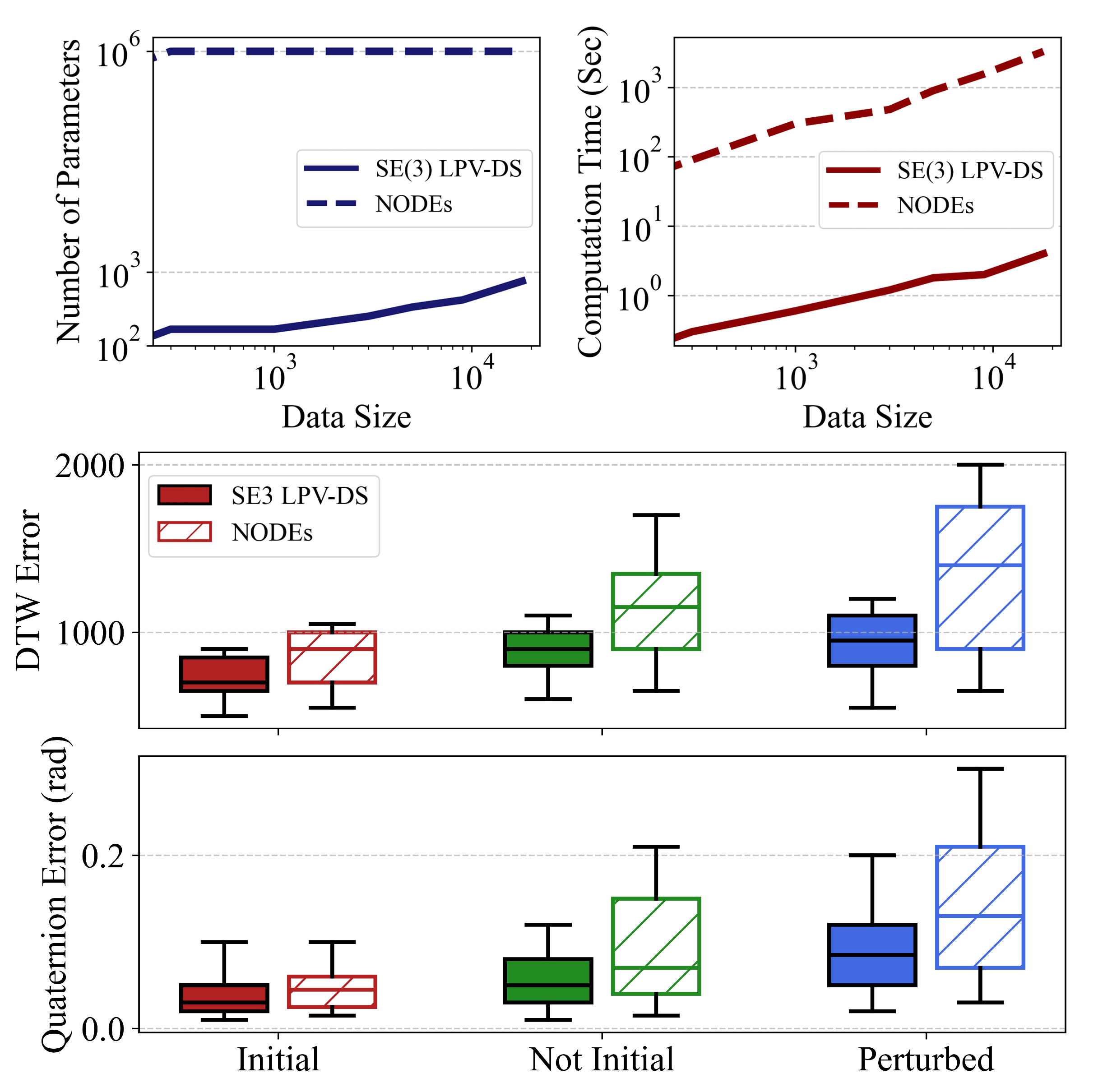}%
\hfill
\caption{Comparison of parameter size and computation time with data size (top), and DTW error and quaternion error (lower is better) over 9 tasks in three scenarios. Solid bars represent our approach, and hatched bars depict the baseline. Bars range from lower quartile to upper quartile, with whiskers indicating extreme values.}
\label{fig:metric}
\vspace{-10pt}
\end{figure}

\textbf{Results: } The model complexity in terms of parameter size for NODEs is pre-defined by the layers of neural networks and remains fixed regardless of trajectory size. By contrast, our approach requires more LTI systems with more parameters to represent an increasing trajectory. \textit{Computation time} for NODEs is largely determined by the iterations as opposed to our approach growing linearly with data size. Nevertheless, our approach is formulated as a semi-definite optimization that can be solved in near real-time, requiring significantly fewer parameters and less time for convergence than NODEs by an order of magnitude as shown in Fig.~\ref{fig:metric}. 

When assessing the \textit{reproduction accuracy}, we report three different scenarios: starting from initial points, starting from unmodeled regions, and reacting to perturbations. We notice that when starting from the provided initial points, both approaches reproduced the trajectory with low mean error and small variance. However, when starting from \textit{outside} the provided initial points, our approach still managed to maintain a comparable accuracy (low error in both position and quaternion) while the baseline experienced a few trials of divergence and instability resulting in larger variance and upper extremes. Similar results occur to the added perturbation: our model, though suffers a slight increase in error, ensures a low variance and prevents any upper outlier. On the other hand, the baseline failed to generalize a stable vector field outside the unmodeled region, leading to worse performance in both position and orientation.

\subsection{Real Robot Experiments}
\begin{figure*}[!htp]
\centering
\includegraphics[width=0.90\linewidth]{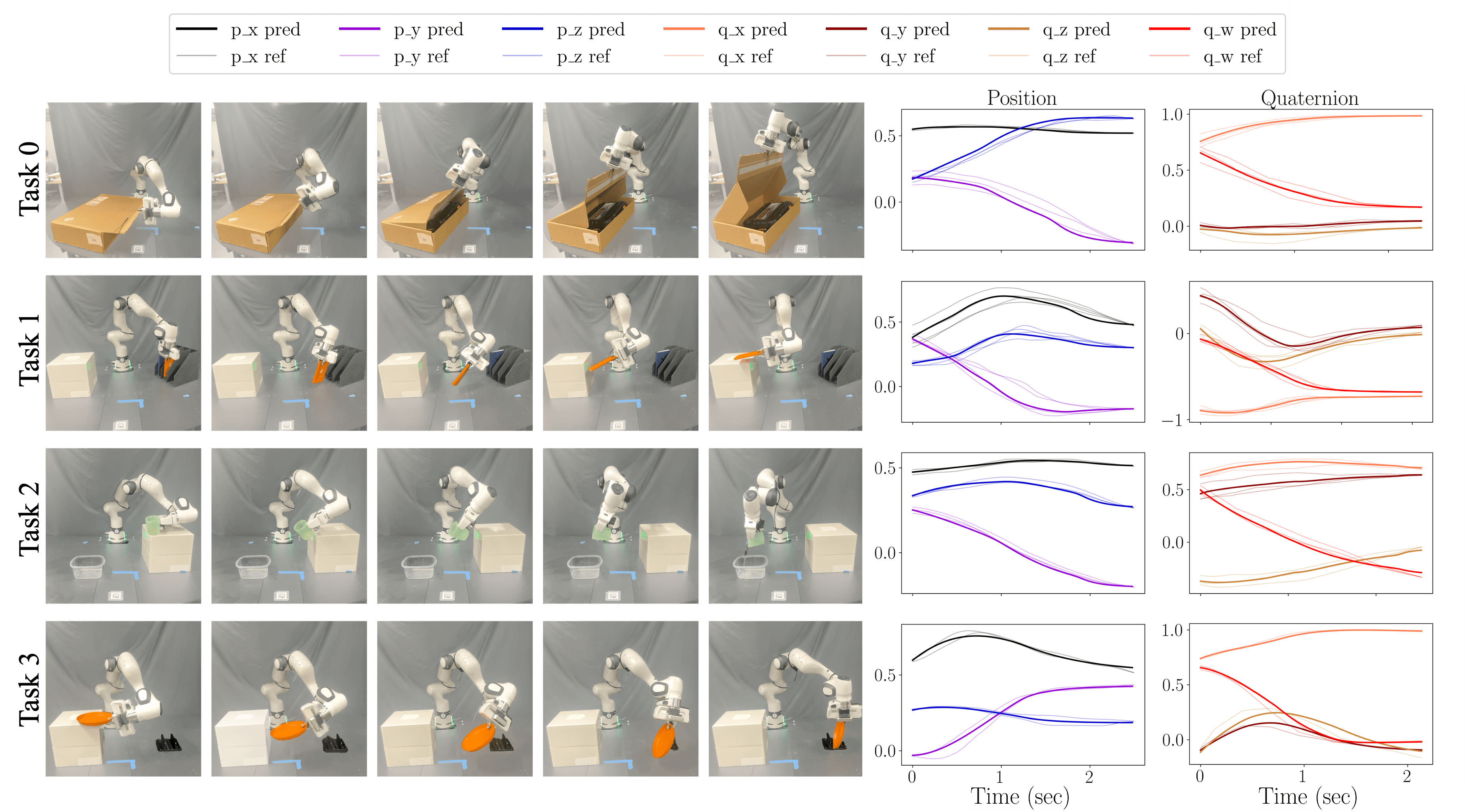}%
\hfill
\caption{A robot trained via SE(3) LPV-DS to perform four different real-world tasks: 0) \textit{box opening}, 1) \textit{book shelving}, 2) \textit{water pouring}, 3) \textit{plate moving}. The training trajectories in which both position and orientation vary over time are provided by kinesthetic demonstrations. For each task, we illustrate the time evolution of the reproduced trajectory --- position and quaternion in their respective coordinates. \label{fig:demo}}
\vspace{-15pt}
\end{figure*}

We validate our approach on four exemplary real-world tasks including \textit{box opening}, \textit{book shelving}, \textit{water pouring} and \textit{plate moving}. Each task contains 3 demonstrations including position and orientation with a total size of approximately 2000 observations. In Figure.~\ref{fig:demo}, we show the sequence of snapshots of a robot successfully performing the four tasks, as well as the evolution of position and quaternion over time. We further illustrate its robustness to perturbations and ability to generalize to the unmodeled region in the supplementary video. 

\section{CONCLUSIONS}
In this paper, we present an extension to the current LPV-DS framework, enabling the learning of trajectory planning/control on orientation and full pose trajectory. Comparing against the baseline methods on real-world tasks, we verified that our approach achieves comparable results on reproduction accuracy and generalization ability while maintaining an efficient model complexity and computation efficiency. However, we note that the underlying assumption of LPV-DS requires no \textit{self-intersecting} trajectory, excluding the broader range of tasks in real life. This leads to the future work where the learned Dynamical System should incorporate higher-order derivative with sequential information.









\bibliographystyle{IEEEtran}
\bibliography{reference}

\end{document}